\documentclass[conference]{IEEEtran}
\IEEEoverridecommandlockouts
\usepackage{cite}
\usepackage{amsmath,amssymb,amsfonts}
\usepackage{algorithmic}
\usepackage{graphicx}
\usepackage{textcomp}
\usepackage{graphicx}
\usepackage{amsmath,amssymb} 
\usepackage{amsmath,amsfonts}
\usepackage{algorithmic}
\usepackage{algorithm}
\usepackage{array}
\usepackage{textcomp}
\usepackage{stfloats}
\usepackage{url}
\usepackage{bbm}
\usepackage{multirow}
\usepackage{verbatim}
\usepackage{threeparttable}
\usepackage{graphicx}
\usepackage{booktabs}
\usepackage{makecell}
\usepackage{diagbox}
\usepackage{arydshln}
\usepackage{color}
\usepackage{cite}
\usepackage{bm}
\usepackage{bbding}
\usepackage{xcolor}
\def\BibTeX{{\rm B\kern-.05em{\sc i\kern-.025em b}\kern-.08em
    T\kern-.1667em\lower.7ex\hbox{E}\kern-.125emX}}
\begin{document}

\title{Efficient Prototype Consistency Learning in Medical Image Segmentation via Joint Uncertainty and Data Augmentation }

\author{\IEEEauthorblockN{Lijian Li}
\IEEEauthorblockA{
\textit{Faculty of Science and Technology} \\
\textit{University of Macau}\\
Macau, China \\
mc35305@umac.mo}
\and
\IEEEauthorblockN{Yuanpeng He}
\IEEEauthorblockA{\textit{Key Laboratory of High Confidence} \\
\textit{Software Technologies (MOE)} \\
\textit{School of Computer Science} \\
\textit{Peking University}\\
Beijng, China \\
heyuanpeng@stu.pku.edu.cn}
\and
\IEEEauthorblockN{Chi-Man Pun}
\IEEEauthorblockA{
\textit{Faculty of Science and Technology} \\
\textit{University of Macau}\\
Macau, China \\
cmpun@umac.mo} 
\and
}
\maketitle

\begin{abstract}
Recently, prototype learning has emerged in semi-supervised medical image segmentation and achieved remarkable performance. However, the scarcity of labeled data limits the expressiveness of prototypes in previous methods, potentially hindering the complete representation of prototypes for class embedding. To overcome this issue, we propose an efficient prototype consistency learning via joint uncertainty quantification and data augmentation (EPCL-JUDA) to enhance the semantic expression of prototypes based on the framework of Mean-Teacher. The concatenation of original and augmented labeled data is fed into student network to generate expressive prototypes. Then, a joint uncertainty quantification method is devised to optimize pseudo-labels and generate reliable prototypes for original and augmented unlabeled data separately. High-quality global prototypes for each class are formed by fusing labeled and unlabeled prototypes, which are utilized to generate prototype-to-features to conduct consistency learning. Notably, a prototype network is proposed to reduce high memory requirements brought by the introduction of augmented data. Extensive experiments on Left Atrium, Pancreas-NIH, Type B Aortic Dissection datasets demonstrate EPCL-JUDA's superiority over previous state-of-the-art approaches, confirming the effectiveness of our framework. The code will be released soon. 
\end{abstract}


\section{Introduction}
Recently, supervised medical image segmentation has achieved remarkable improvements by introducing deep learning methods. However, the widespread application of such techniques in real medical diagnosis is continually hindered by the scarcity of labeled data. Thus, researchers have proposed the concept of semi-supervised medical image segmentation (SS-MIS) to reduce the dependence of models on abundant manual annotations, which require a significant amount of time and labor. SS-MIS methods are capable of achieving relatively great performance by extracting precious information from unlabeled data to assist model in training with a small amount of annotated data.

There are three common categories of semi-supervised learning techniques utilized in SS-MIS methods, including consistency-based, pseudo-label-based, and prototype-based learning approaches. As for pseudo-label-based methods \cite{DBLP:conf/cvpr/KwonK22, DBLP:conf/miccai/BaiOSSRTGKMR17, DBLP:conf/nips/ZhangWHWWOS21, DBLP:conf/cvpr/ChenYZ021, he2024uncertainty}, they aim to generate pseudo-labels for unlabeled data to provide auxiliary supervision similar to that of annotated data. Some studies introduce uncertainty estimation methods like Monte Carlo dropout \cite{yu2019uncertainty} and ensemble-based methods \cite{shi2021inconsistency, he2025co} to mitigate interference information contained in unlabeled data so as to generate more reliable pseudo-labels. Besides, researchers also quantify the level of uncertainty and set a threshold to select high-confidence pseudo-labels by utilizing some measurements like information entropy. Xiang et al. \cite{xiang2022fussnet} considers both epistemic uncertainty (EU) and aleatoric uncertainty (AU) to generate a better uncertainty mask to promote model learning. Consistency-based methods \cite{DBLP:conf/nips/BachmanAP14, DBLP:conf/nips/SajjadiJT16, DBLP:conf/miccai/BortsovaDHKB19, DBLP:conf/icml/XuSYQLSLJ21, DBLP:journals/corr/abs-2001-04647, DBLP:conf/bmvc/FrenchLAMF20, DBLP:conf/aaai/LuoCSW21, he2024generalized} comply with a learning pattern that guides model to generate consistent predictions for both raw data and enhanced or interfered data. A unified consistency loss is adopted to optimize model learning. Chen et al. \cite{DBLP:conf/cvpr/ChenYZ021} proposes a learning strategy named cross pseudo supervision (CPS) to enable pseudo labels generated by two perturbed segmentation networks to supervise mutually. Luo et al. \cite{DBLP:conf/aaai/LuoCSW21} introduces a dual-task consistency regularization (DTC) to enforce consistency between pixel-level and geometry-level segmentation maps for labeled and unlabeled data. However, slight or inappropriate perturbations may provide erroneous supervisory signals so as to produce suboptimal segmentation performance. In regard to prototype learning methods \cite{DBLP:conf/cvpr/WangWSFLJWZL22, DBLP:journals/pami/WuFHHMZ23, DBLP:conf/cvpr/Zhang0Z0WW21, bi2025multi, he2024epl}, they are intended to impose a consistency constraint between prototypes generated by feature matching operations and corresponding model's predictions to improve segmentation performance. Xu et al. \cite{DBLP:journals/titb/XuWLYY000T22} devises a cyclic prototypical consistency learning framework (CPCL) that is composed of a labeled-to-unlabeled (L2U) prototypical forward process and an unlabeled-to-labeled (U2L) backward process. However, current prototype learning methods always generate prototypes for labeled and unlabeled data separately and fuse two kinds of prototypes to obtain more comprehensive prototypes. They are limited by the quantity and quality of prototypes so that the representation capability of global prototypes will be impaired. In response to the insufficient amount of annotated data, researchers usually use interpolation-based semi-supervised learning methods to solve this problem. For example, CutMix \cite{yun2019cutmix}, also known as Copy-Paste (CP), achieves data augmentation by copying image crops from labeled data as foreground onto another image. It has been verified that the predictions of augmented data can provide useful information to assist in training the original annotated data.

Briefly, previous prototype-based methods are hindered by the limited quantity and low quality of prototypes. To address this problem, we propose an efficient prototype consistency learning method via joint uncertainty quantification and data augmentation (EPCL-JUDA) to generate high-quality global prototypes for consistency learning based on the Mean-Teacher framework \cite{li2024efficient}. Firstly, we enhance the labeled data using augmentation techniques like CutMix \cite{yun2019cutmix} and concatenate it with the original labeled data. The concatenated labeled data is then fed into the student network, generating more expressive prototypes using the masked average pooling approach \cite{wang2019panet}. Secondly, we process the original and augmented unlabeled data separately utilizing a joint uncertainty quantification method, combining entropy and distributional uncertainty to reduce the interference introduced by data augmentation operations and further screen more reliable pseudo-labels which are used as masks to generate two distinct unlabeled prototypes. The fusion of these prototypes generates expressive prototypes for unlabeled data. Comprehensive global prototypes containing abundant common semantic information are formed by fusing labeled prototypes and unlabeled prototypes, which have enhanced capabilities to represent class embeddings. Notably, with the introduction of augmented data, directly generating prototypes using features in the decoder, as in previous work \cite{lu2023upcol}, results in significant memory consumption. Therefore, we design a prototype network to generate prototypes, effectively reducing the VRAM required for prototype generation and further improving segmentation performance to some extent. Finally, the prototype-to-feature similarity maps for labeled and unlabeled data are regarded as novel segmentation masks and are used to conduct consistency learning with labels and reliable pseudo-labels.

In sum, the main contributions of this work are summarized as follows:
\begin{itemize}
    \item[1)] We propose an enhanced prototype consistency learning via joint uncertainty quantification and data augmentation to generate high-quality prototypes to better represent class embeddings, which increases the number of samples the process of prototype generation can consider to improve the quality of global prototypes utilized to generate prototype-to-feature similarity maps to conduct consistency learning .

    \item[2)] A prototype network is devised to reduce significant memory consumption for prototype generation resulting from prototype generation method and the introduction of augmented data.

    \item[3)] The results of extensive experiments conducted on Left Atrium, Pancreas-CT, Aortic Dissection datasets verify that the proposed EPCL-JUDA framework outperforms previous SOTA semi-supervised SSL methods.
\end{itemize}

\section{Related works}
\subsection{Semi-supervised Medical Image Segmentation}
Recently, numerous semi-supervised learning methods have been introduced into the field of medical image segmentation and achieved quite considerable breakthroughs. The existing semi-supervised medical image segmentation (SS-MIS) methods generally utilize an encoder-decoder segmentation network like V-Net as their backbone and mainly focus on devising an excellent learning strategy \cite{he2024mutual}. Consistency learning is one of the most commonly used strategies. Most consistency learning-based methods adopt the structure of Mean Teacher and design various regularization methods to maintain the consistency of pseudo-labels generated by teacher and student networks. For example, Yu et al. \cite{yu2019uncertainty} applies an uncertainty-guided learning strategy based on the Mean Teacher framework to enable the student network to learn more reliable targets, with Monte Carlo Dropout used for uncertainty estimation. Related ideas have been extended by works incorporating evidential theory to enhance robustness in label propagation and regularization \cite{he2024uncertainty}. Wu et al. \cite{DBLP:conf/miccai/WuXGCZ21} devise a mutual consistency network (MC-Net) that constrains two parallel decoder predictions, mitigating ambiguity in unlabeled data. Xia et al. \cite{DBLP:journals/mia/XiaYYLCYZXYR20} attempts to enforce the consistency of predictions for multi-view unlabeled data. Furthermore, Ouali et al. \cite{ouali2020semi} proposes a cross-consistency strategy by applying perturbations to different decoder branches, while Luo et al. \cite{luo2021efficient} develops a multi-scale uncertainty rectification strategy. Recently, some studies have attempted to improve evidence combination for label reliability under weak supervision using conflict-aware fusion mechanisms \cite{he2022new, he2021conflicting}.

Prototype alignment based on feature matching has become another popular direction in SS-MIS, which leverages class-level semantic representation to enhance distribution regularity of features. Wu et al. \cite{wu2022exploring} aligns category-specific features with global prototypes to improve intra-class compactness. Zhang et al. \cite{DBLP:conf/cvpr/Zhang0Z0WW21} proposes a self-correcting strategy using prototype distances to adjust pseudo-labels, while Zhang et al. \cite{zhang2020sg} utilizes masked average pooling and cosine similarity to perform fine-grained feature-to-prototype matching. These methods have been enhanced by uncertainty-aware prototype representation techniques, such as evidential prototype learning \cite{he2024epl} and multi-prototype refinement \cite{bi2025multi}. Furthermore, certain prototype-based strategies incorporate ordinal entropy \cite{he2022ordinal} and belief-based uncertainty to optimize alignment robustness.

Adversarial learning has also shown promise in semi-supervised segmentation. Peiris et al. \cite{peiris2021duo} combines adversarial and multi-view learning for pseudo-mask refinement. Wang et al. \cite{wang2023cat} introduces constraint-driven adversarial training that enforces anatomical plausibility. Complementary efforts explore hybrid attention \cite{he2024generalized} and entropy optimization \cite{li2025adaptive} in label-uncertain scenarios.

\subsection{Interpolation-based Semi-supervised Learning}

To address the limited availability of labeled data, interpolation-based methods have been widely adopted in semi-supervised semantic segmentation. Mixup \cite{zhang2017mixup} and CutMix \cite{yun2019cutmix} are fundamental strategies that blend image or region information to create diverse training samples. For example, Tu et al. \cite{tu2022guidedmix} uses mixup to propagate label information into unlabeled data, while Dvornik et al. \cite{dvornik2018modeling} uses object context for realistic placement of augmented content.

Bai et al. \cite{bai2023bidirectional} proposes a bidirectional copy-paste method that combines labeled and unlabeled crops to generate more semantically complete examples. Berthelot et al. \cite{berthelot2019mixmatch} incrementally incorporates high-confidence pseudo-labeled data to expand the annotated set. Building upon these concepts, EPCL-JUDA \cite{li2024efficient} combines CutMix augmentation and prototype consistency learning with uncertainty modeling, resulting in effective global prototypes and refined pseudo-labels.

Recent works also incorporate quantification of distributional ambiguity into interpolation strategies. For instance, the use of ordinal belief entropy \cite{he2023ordinal} and quantum mass functions \cite{he2023tdqmf} helps capture soft uncertainty signals in mixed training samples. These ideas contribute to more reliable decision boundaries, particularly in cross-domain or limited-sample settings. The integration of such uncertainty modeling strategies into prototype frameworks further enhances model generalizability under weak supervision \cite{he2025co}. 

Additionally, cross-institutional medical collaboration scenarios, where annotation distribution is highly uneven, benefit from vertical federated frameworks such as UniTrans \cite{huang2025unitrans}, which aim to mitigate privacy and annotation imbalance constraints in semi-supervised training environments.

\section{Methodology}
\subsection{Problem Definition}
\begin{figure*}
    \centering
    \includegraphics[scale=0.85]{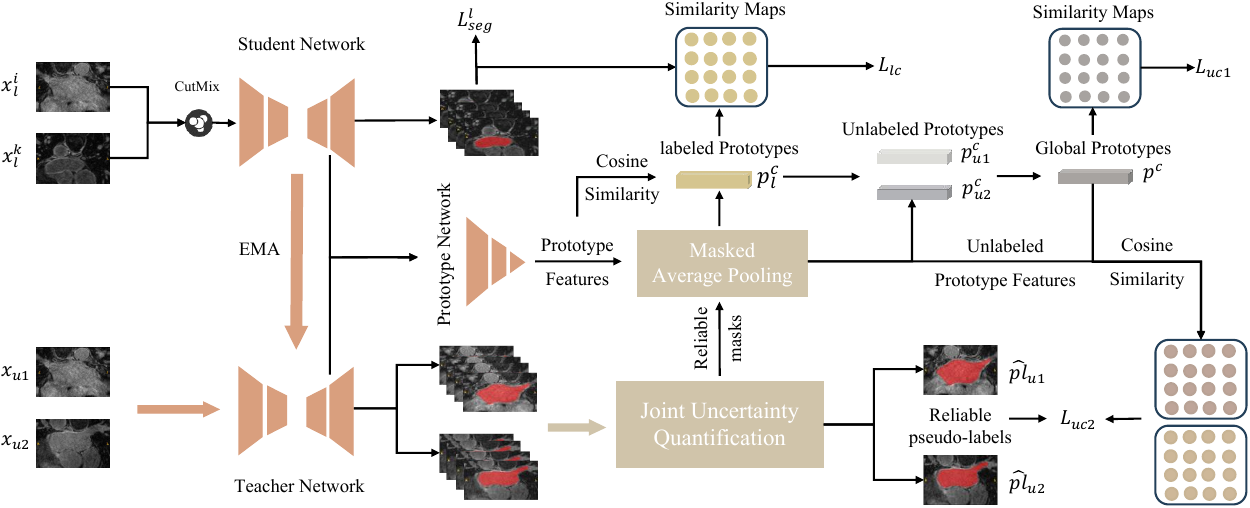}
    \caption{The overview of the proposed EPCL-JUDA framework. For training process, labeled data is augmented using techniques like CutMix and combined with original data, fed into the teacher network to generate expressive prototypes $p^c_l$. And, a prototype network is introduced to generate prototype features, which effectively reduces memory consumption and improves segmentation performance. Additionally, a Joint uncertainty quantification, combining entropy, variance, and mean, is employed to optimize pseudo-labels and generate reliable prototypes $p^c_{u1}$ and $p^c_{u2}$ for original and augmented unlabeled data. These prototypes are fused to form high-quality global prototypes $p^c$, which are used to generate multiple prototype-to-feature similarity maps which are utilized to conduct consistency learning with labels and pseudo-labels, with different prototype features.}
    \label{structure}
\end{figure*}

In the task of semi-supervised medical image segmentation, we have a whole dataset $\mathbb{D}$ including an annotated subset and unannotated subset, which are defined as $\mathbb{D}_l, \mathbb{D}_u$ respectively. The annotated subset consists of $N$ images and corresponding labels denoted as $\mathbb{D}_l = \{(x^a, y^a)\}_{a=1}^{\mathcal{B}}$, and the unlabeled subset only contains $M$ images denoted as $\mathbb{D}_u = \{(x^a)\}_{a=\mathcal{B}+1}^{2\mathcal{B}}$. Besides, the dimensions of each image are $H \times W \times D$ which represent height, width and depth. Therefore, we denote a single image as $x^a \in \mathbb{R}^{H \times W \times D}$. For each label, it is defined as $y^a \in \{0,1,...,C-1\}^{H \times W \times D}$. The batch size for both labeled and unlabeled is 2, which is denoted as $\mathcal{B}$.The proposed EPCL-JUDA framework is composed of a Mean Teacher, prototype network and joint uncertainty quantification, whose overview structure is illustrated in Fig. \ref{structure}. The student network is optimized by optimizers like stochastic gradient descent (SGD). Then, the parameters of teacher network are updated by exponential moving average (EMA) of student network.
\subsection{Supervised Loss}
To mitigate the uncertainty of predictions for labeled data that model brings, we employ four different loss functions, such as focal loss, Dice loss, cross entropy loss and IoU loss, to process different outputs from the same group of classifiers. Additionally, unlike \cite{xiang2022fussnet}, we also utilize the cross-entropy loss function to process the average of four distinct predictions to further eliminate disparities between the results of the four classifiers. Therefore, the supervised loss for labeled data is the sum of the average of the four different losses and the fused cross-entropy loss, which is defined as follows:
\begin{equation}
    L_{seg}^l = (L_{ce}^l + L_{dice}^l + L_{focal}^l + L_{IoU}^l) / 4 + L^l_f 
\end{equation}
where $L_{ce}, L_{dice}, L_{focal}, L_{IoU}$ represent cross entropy loss, dice loss, focal loss and IoU loss for labeled data respectively and $L_{f}$ denotes the fused cross entropy loss.

\subsection{Joint Uncertainty Quantification}
The pseudo-label for the $p$-th voxel in the $a$-th unlabeled image is denoted as $pl_{u,a}^p \in \mathcal{R}^C$. Similarly, the average of multiple different predictions $\bar{P}_u$ for unlabeled images will be utilized to generate pseudo labels. To progressively improve the reliability of pseudo-labels, we introduce a comprehensive uncertainty measurement named Joint Uncertainty Quantification (JUQ), which combines entropy and distributional uncertainty, to accurately assess each voxel's uncertainty. Traditional methods for assessing uncertainty often rely solely on entropy, which measures the randomness or unpredictability of the prediction distribution. According to the properties of entropy, the lower the entropy of a voxel, the lower its uncertainty, implying higher reliability. The entropy-based uncertainty for the $p$-th voxel is defined as follows:
\begin{equation}
    U_{entropy}(pl_{u,a}^p) = -\sum_{c=0}^{C-1}pl_{u,a}^p(c)\log pl_{u,a}^p(c)
\end{equation}
While entropy provides valuable insights into the uncertainty of predictions, entropy is highly sensitive to noise. When processing augmented unlabeled data, relying solely on entropy for uncertainty estimation may cause over-sensitivity. To address this limitation, we propose the Joint Uncertainty Quantification approach, which combines entropy with distributional uncertainty to offer a more holistic assessment of uncertainty. JUQ is capable of balancing noise impact and enhancing robustness in evaluating uncertainty, particularly for noisy, complex data. Besides, JUQ is also capable of strengthening model's ability to distinguish between high and low uncertainty samples, further improving the reliability of pseudo-labels. For each voxel, the combined uncertainty $JUQ$ is defined as:
\begin{align}
&U_{\text{Dis-norm}}(\bar{P}_{u,a}^p) = \exp\left(-\frac{\text{Var}(\bar{P}_{u,a}^p)}{\sum_{p=1}^{H \times W \times D}\text{Var}(\bar{P}_{u,a}^p)}\right)\\
&U_{\text{entropy-norm}}(pl_{u,a}^p) = 1 - \frac{U(pl_{u,a}^p)}{\sum_{p=1}^{H \times W \times D}U(pl_{u,a}^p)} \\
&JUQ = U_{\text{Dis-norm}}(\bar{P}_{u,a}^p) \cdot U_{\text{entropy-norm}}(pl_{u,a}^p)
\end{align}
where Var$(\bar{P}_{u,a}^p)$ denotes the variance of mean predictions for the $p$-th voxel. With the comprehensive reliability map generated from JUQ, it is regarded as weight assigned to each voxel. Therefore, the reliable pseudo labels for unlabeled data can be obtained by multiplying the reliability map with raw pseudo-labels. The process can be defined as follows:
\begin{equation}
\hat{pl}_{u,a}^p = \frac{1}{H\times W\times D} (1 - \frac{JUQ}{\sum_{p=1}^{H\times W \times D}JUQ})\otimes pl_{u,a}^p
\end{equation}
where $\hat{pl}_{u,a}^p$ represents reliable pseudo-label for $p$-th voxel, and $\otimes$ denotes element-wise multiplication operation.

\subsection{Enhanced Prototype Learning}
\textbf{Prototype Generation.} In previous work \cite{lu2023upcol}, the features from the $k$-rd layer decoder are upsampled to the same size as segmentation labels and then are utilized to generate prototypes. However, such a method requires a significant amount of GPU memory to process high-dimensional decoder features, especially after incorporating additional augmented data. Differently, we devise a prototype network consisting of three 3D convolutional layers is to gradually processes and reduces feature dimensions to the number of prototypes or classes, through multiple convolutional layers. Then, upsampling the low-dimensional features generated by prototype network significantly reduces the memory demand of model and improve the performance for a certain extent. A detailed analysis will be provided in ablation experiments \ref{memory}. Besides, due to the limited annotated data, the annotated prototypes cannot fully express the semantic embeddings of specific classes. To address the problem, we apply some data augmentation techniques like cutmix \cite{yun2019cutmix} to enrich the annotated samples that can be considered in the prototype generation process. Here, we set the prototype features produced by the prototype network to be $f_{a}^{l, p}$ for the concatenation of original and augmented labeled data. Referencing \cite{wang2019panet}, we directly utilize labels to mask the feature maps and then adopt masked average pooling to obtain the prototype for each class. The labeled prototype for class $c$ is defined as follows:
\begin{equation}
    p^c_l = \frac{2}{3\mathcal{B}}\sum_{a=1}^{\frac{3\mathcal{B}}{2}} \frac{\sum_{p=1}^{H\times W \times D} f_{a}^{l, p} \mathbbm{1}[y^a_p=c]}{\sum_{p=1}^{H\times W \times D}\mathbbm{1}[y^a_p=c]}
\end{equation}
With more labeled data for reference, the labeled prototypes will have a stronger semantic representation capability. Different from the generation of labeled prototypes,we utilize the proposed joint uncertainty quantification measure the uncertainty of each voxel to obtain reliability pseudo-labels $\hat{pl}_{u1,a}^p$ and $\hat{pl}_{u2,a}^p$ as well as two reliable maps for original and augmented unlabeled data separately. Before conducting the operation of masked attention pooling, the prototype features $f_{a}^{u1, p}$ of original unlabeled data are weighted by the augmented reliability  map. The original unlabeled prototype for class $c$ is formulated as follows:
\begin{equation}
    p^c_{u1} = \frac{1}{\mathcal{B}}\sum_{a=1}^\mathcal{B}  \frac{(1 - \frac{JUQ}{\sum_{p=1}^{H\times W \times D} JUQ})}{H\times W\times D}\frac{\sum_{p=1}^{H\times W \times D}   f_{a}^{u1, p}  \mathbbm{1}[y^a_p=c]}{\sum_{p=1}^{H\times W \times D}\mathbbm{1}[y^a_p=c]}
\end{equation}
For the augmented unlabeled data, the optimized pseudo-label $\hat{pl}_{u2,a}^p$ is utilized to provide a credible mask to conduct masked average pooling to generate prototypes. The augmented unlabeled prototype for class c is defined as follows:
\begin{equation}
    p^c_{u2} = \frac{1}{\mathcal{B}}\sum_{a=1}^\mathcal{B}  \frac{(1 - \frac{JUQ}{\sum_{p=1}^{H\times W \times D} JUQ})}{H\times W\times D}\frac{\sum_{p=1}^{H\times W \times D}   f_{a}^{u2, p}  \mathbbm{1}[y^a_p=c]}{\sum_{p=1}^{H\times W \times D}\mathbbm{1}[y^a_p=c]}
\end{equation}
where $f_{a}^{u2, p}$ denotes the prototype features for augmented unlabeled data. 

\textbf{Prototype Fusion.}
With the augmented unlabeled prototypes $p^c_{u2}$, they are regarded as additional semantic knowledge to enhance the semantic information of original unlabeled prototypes $p^c_{u1}$. Expressive unlabeled prototypes $p^c_u$ will be generated by fusing original and augmented unlabeled prototypes, which have better capabilities to express class embeddings. The fusion process is defined as follows:
\begin{equation}
    p^c_{u} = \lambda_1 p^c_l + \lambda_2 p^c_m
\end{equation}
where $\lambda_1$ and $\lambda_2$ are the coefficients to adjust the fusion ratio of original and augmented unlabeled prototypes. 

Finally, we fuse the optimized labeled and unlabeled prototypes to generate high-quality global prototypes $p^c$, which further bridge the distribution gap between labeled and unlabeled data and have better expression ability for class embeddings. We adopt Temporal Ensembling approach that is capable of gradually improving the quality of predictions and prototypes during the training process to progressively update the fusion percentage of distinct prototypes. Time-dependent Gaussian warming up function \cite{tarvainen2017mean} is a prevalent technique that includes a vital parameter $\lambda_{con}$ ranging from 0 to 1. As the training progresses, the coefficients for labeled and unlabeled prototypes enable model to continually focus on the expressive unlabeled prototypes while maintaining labeled prototypes as the capital information sources.
\begin{equation}
    p^c = ((2 - \lambda_{con})p^c_{l} + \lambda_{con}p^c_{u}) / 2
\end{equation}

\subsection{Loss functions}
With the global prototypes, the next step is to build relationships between prototypes and features from teacher and student networks. We utilize cosine distance to measure the similarity \cite{zhang2020sg}. Then, the likelihood of voxels in each class is approximated using feature-to-prototype similarity maps.
To conduct consistency learning, we utilize cross entropy loss function to narrow the distance between prototype features for labeled data and labels. For unlabeled data, original, augmented and their average prototype features ($f_{a}^{u1, p}, f_{a}^{u2, p}, f_{a}^{u, p}$) are regarded as beneficial information and are expected to close to pseudo labels $\hat{pl}_{u2,a}^p$ of augmented unlabeled data to improve their qualities. The consistency losses based on prototypes for labeled and unlabeled data are defined as follows:
\begin{equation}
    L_{lc} = L_{ce}(s_{l, a}, y_a),  L_{uc1} = \sum_{p=1}^{H\times W \times D} L_{ce}(s_{u, a}^p, \hat{pl}_{u2, a}^p)
\end{equation}
\begin{equation}
    L_{uc2} = \sum_{p=1}^{H\times W \times D}L_{ce}(s_{u1, a}^p, \hat{pl}_{u2, a}^p) + L_{ce}(s_{u2, a}^p, \hat{pl}_{u2, a}^p)
\end{equation}
where $s_a^l$ denotes the similarity maps between global prototypes and labeled prototype features, and $s_{u, a}, s_{u1, a}, s_{u2, a}$ denote the similarity maps between prototypes and multiple unlabeled prototype features. Finally, the total loss is combined with all supervised losses and two consistency losses, which is shown in below equation:
\begin{equation}
    L = L_{seg} + L_{lc} + \lambda_{con}(L_{uc1} + L_{uc2})
\end{equation}
\section{Experiments}
\subsection{Datasets and Metrics}
Three public datasets are utilized to evaluate the performance of our model, including the left atrium (LA) dataset \cite{DBLP:journals/mia/XiongXHHBZVRMYH21}, Pancreas-NIH \cite{roth2015deeporgan} and a multi-center dataset for type B aortic dissection (TBAD) \cite{yao2021imagetbad}. For all datasets, we normalize the voxel intensities to zero mean and unit variance.

\textbf{LA dataset} \cite{DBLP:journals/mia/XiongXHHBZVRMYH21} is comprised of 100 3D gadolinium-enhanced magnetic resonance (MR) imaging volumes, each of which has identical spatial resolution of $0.625 \times 0.625 \times 0.625$\(mm^3\) and distinct dimensions. The dataset is methodically partitioned to allocate 80 samples for the training set and 20 samples for the validation set. For training, samples in training set are randomly cropped into $112 \times 112 \times 80$ patches. For inference, a sliding window of the same dimensions is employed, along with a stride of $18 \times 18 \times 4$, to generate the final segmentation outcomes. 

\textbf{Pancreas-NIH dataset} \cite{roth2015deeporgan} consists of 82 contrast-enhanced abdominal CT volumes with manual labels. For fair comparison, we randomly crop the training volumes to $96 \times96 \times 96$ and the stride is $16\times 16 \times 16$ at inference following CoraNet \cite{shi2021inconsistency}.

\textbf{TBAD dataset} \cite{yao2021imagetbad} consists of 124 computed tomography angiography (CTA) scans having three annotations (whole aorta, true lumen (TL), and false lumen (FL)). In the dataset, 100 and 24 scans are utilized for training and testing, respectively. Besides, for the preprocessing of data, we utilize the same techniques as in \cite{lu2023upcol}.

\textbf{Metrics.} We evaluate our model with four common metrics: Dice coefficient (Dice), Jaccard Index (Jac), 95\% Hausdorff Distance (95HD), and Average Symmetric Surface Distance (ASD). Two former and latter metrics are regionally sensitive and edge-sensitive, respectively. 
\begin{table}[htbp]\small
\renewcommand{\arraystretch}{1.1}
\centering
\caption{Experimental Results Comparison on the LA Dataset}
  \setlength{\tabcolsep}{0.8mm}{\begin{tabular}{c|cc|cccc}
    \hline
    \multirow{2}*{Method} &\multicolumn{2}{c|}{Scans Used} &\multicolumn{4}{c}{Metrics}\\
    \cline{2-7}
    {}&{Labeled} & {Unlabled} & Dice$\uparrow$ & Jaccard$\uparrow$ & 95HD$\downarrow$ & ASD$\downarrow$ \\
    \hline
    UA-MT & \multirow{8}*{8(10\%)} &\multirow{8}*{72(90\%)} & 87.79 &78.39 &8.68& 2.12 \\
    SASSNet  &{}&{}& 87.54 &78.05 &9.84 &2.59 \\
    DTC  &{}&{}& 87.51 &78.17 &8.23 &2.36 \\
    URPC &{}&{}& 86.92 &77.03 &11.13 &2.28 \\
    MC-Net &{}&{}& 87.62 &78.25 &10.03 &1.82 \\
    SS-Net &{}&{}& 88.55 &79.62 &7.49 &1.90 \\
    BCP &{}&{} &89.62& 81.31& 6.81 &1.76 \\
    Co-BioNet &{}&{}&  89.20 & 80.68 & 6.44 & 1.90 \\
    UPCoL &{}&{}& - & - & - & -\\
    \hline
    EPCL-EN&\multirow{2}*{8(10\%)} &\multirow{2}*{72(90\%)} & 91.23 & 83.17 & 5.97 & 1.81\\
    EPCL-JUQ&{}&{} & \textbf{91.52} & \textbf{83.76} & \textbf{5.49} & \textbf{1.73}\\
    \hline
    UA-MT & \multirow{8}*{16(20\%)} &\multirow{8}*{64(80\%)} & 88.88 & 80.21 & 7.32 & 2.26 \\
    SASSNet  &{}&{}& 89.54 & 81.24 & 8.24 & 1.99 \\
    DTC  &{}&{}& 89.42 & 80.98 & 7.32 & 2.10 \\
    URPC &{}&{}& 88.43 & 81.15 & 8.21 & 2.35 \\
    MC-Net &{}&{}& 90.12 & 82.12 & 11.28 & 2.30 \\
    SS-Net &{}&{}& 89.25 & 81.62 & 6.45 & 1.80 \\
    BCP &{}&{} &90.34 & 82.50 & 6.75 & 1.77 \\
    Co-BioNet &{}&{}& 91.26 & 83.99 & 5.17 & 1.64 \\
    UPCoL &{}&{}& 91.69 & 84.69 & 4.87 &1.56 \\
    \hline
    EPCL-EN&\multirow{2}*{16(20\%)} &\multirow{2}*{64(80\%)} & 91.88 & 84.86 & 4.78 & 1.66\\
    EPCL-JUQ&{}&{} & \textbf{92.24} & \textbf{85.57} & \textbf{4.63} & \textbf{1.47}\\
    \hline
  \end{tabular}}
  \label{LA}
\end{table}
\subsection{Implementation details}
We employ V-Net \cite{milletari2016v} as backbone. The student network is trained for 14k iterations and optimized by an Adam optimizer with a learning rate of 0.001, while parameters of teacher network are updated using exponential moving average (EMA). The batch size is set to 4, containing 2 annotated and unannotated samples. During the training stage, we augment the TBAD dataset with same approaches like rotation following \cite{lu2023upcol} and use 3-fold cross-validation for TBAD dataset and 5-fold cross-validation for LA and Pancreas-NIH datasets. Besides, the fusion coefficients for different unlabeled prototypes $\lambda_1$ and $\lambda_2$ are equal to 1. The optimal number of prototype network is 3. And the default loss function utilized in consistency learning is cross entropy. CutMix is set to be default data augmentation technique for labeled data. The whole EPCL-JUDA framework is implemented by PyTorch and trained with an RTX 4090 GPU.
\begin{figure}
    \centering
    \includegraphics[scale=0.29]{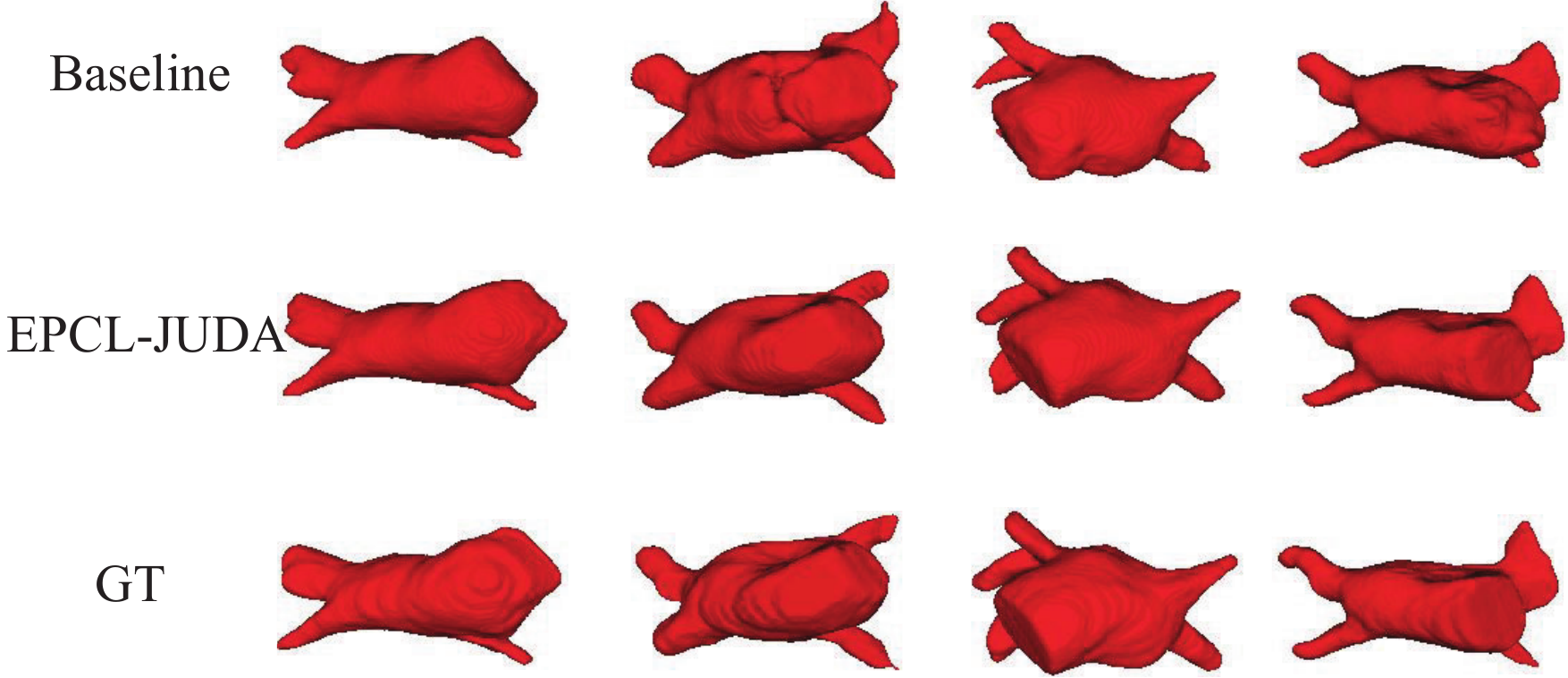}
    \caption{The visualizations of experimental results on Left Atrium dataset. EPCL-JUDA denotes the proposed mixed prototypes consistency learning. GT denotes the ground truth labels. Baseline represents the method of URPC. ITK-SNAP \cite{py06nimg} is the tool to visualize 3D medical image.}
    \label{vis}
\end{figure}
\subsection{Discussions}
\textbf{Results on Left Atrium, Pancreas-NIH, and Aortic Dissection Datasets.}
The models compared are UA-MT \cite{DBLP:conf/miccai/YuWLFH19}, SASSNet \cite{DBLP:conf/miccai/LiZH20}, DTC \cite{DBLP:conf/aaai/LuoCSW21}, URPC \cite{luo2022semi}, MC-Net \cite{DBLP:conf/miccai/WuXGCZ21}, SS-Net \cite{wu2022exploring}, FUSSNet \cite{xiang2022fussnet}, Co-BioNet \cite{peiris2023uncertainty}, BCP \cite{bai2023bidirectional}, UPCoL \cite{lu2023upcol}. Table \ref{LA} presents a comparison of various state-of-the-art (SOTA) models on the Left Atrium dataset, focusing on their performance with labeled ratios of 10\% and 20\%. Notably, across all metrics (Dice, Jaccard, 95HD, ASD), EPCL-JUQ consistently surpasses the performance of previous SOTA models across all datasets. EPCL-EN denotes that replacing JUA in the framework of EPCL-JUA as entropy, whose performance is second to EPCL-JUQ. 
For both 10\% and 20\% labeled ratio, EPCL-JUQ continues to exhibit significant improvement with a Dice score of 92.24, a Jaccard index of 85.57, and the lowest 95HD (4.63) and ASD (1.47), indicating its robust performance. Table \ref{Pancreas} presents a comparison of various SOTA models on Pancreas-NIH dataset under the scenery of 10\% and 20\% labeled ratios. 
Similarly, with a 20\% labeled ratio, EPCL-JUQ continues to achieve best results with a Dice score of 84.45, a Jaccard index of 73.61. In the Aortic Dissection dataset comparison, as illustrated in Table \ref{1}, EPCL-JUDA framework also excels, showing remarkable performance improvements with 10\% and 20\% labeled data. Under the scenery of 20\% labeled data, EPCL-JUDA achieves the highest Dice (up to 80.89\%), Jaccard (up to 69.14\%), and significantly lower 95HD (down to 3.45) and ASD (down to 0.71) metrics, showing robust capability of capturing sophisticated geometric information like the vessel walls between True Lumen (TL) and False Lumen (FL). 
\begin{table}[htbp]\small
\renewcommand{\arraystretch}{1.1}
\centering
\caption{Experimental Results Comparison on the Pancreas-NIH Dataset}
  \setlength{\tabcolsep}{0.8mm}{\begin{tabular}{c|cc|cccc}
    \hline
    \multirow{2}*{Method} &\multicolumn{2}{c|}{Scans Used} &\multicolumn{4}{c}{Metrics}\\
    \cline{2-7}
    {}&{Labeled} & {Unlabled} & Dice$\uparrow$ & Jaccard$\uparrow$ & 95HD$\downarrow$ & ASD$\downarrow$ \\
    \hline
    UA-MT & \multirow{8}*{6(10\%)} &\multirow{8}*{56(90\%)} & 77.26 &63.82 &11.90 &3.06 \\
    SASSNet  &{}&{}& 77.66 &64.08 &10.93 &3.05 \\
    DTC  &{}&{}& 78.27 &64.75 &8.36 &2.25 \\
    URPC &{}&{}& 64.73 &49.62 &21.90 &7.73 \\
    MC-Net &{}&{}& 69.07 &54.36 &14.53 &\textbf{2.28} \\
    SS-Net &{}&{}& 67.40 &53.06 &20.15 &3.47 \\
    UPCoL&{}&{} &-& -& - &- \\
    BCP &{}&{} &81.54& 69.29& 12.21 &3.80 \\
    Co-BioNet &{}&{}&  82.49 & 67.88 & 6.51 & 3.26 \\
    \hline
    EPCL-EN&\multirow{2}*{6(10\%)} &\multirow{2}*{56(90\%)} & 82.53 & 69.87 & 6.24 & 3.01\\
    EPCL-JUQ&{}&{} & \textbf{83.16} & \textbf{70.49} & \textbf{5.97} & 2.85\\
    \hline
    UA-MT & \multirow{8}*{12(20\%)} &\multirow{8}*{50(80\%)} & 77.26 &63.82 &11.90 &3.06 \\
    SASSNet  &{}&{}& 77.66 &64.08 &10.93 &3.05 \\
    DTC  &{}&{}& 78.27 &64.75 &8.36 &2.25 \\
    URPC &{}&{}& 79.09 & 65.99 & 11.68 & 3.31 \\
    MC-Net &{}&{}& 78.17 & 65.22 &6.90 & 1.55 \\
    SS-Net &{}&{}& 79.74 & 65.42 & 12.44 & 2.69 \\
    UPCoL &{}&{}& 81.78 & 69.66 & \textbf{3.78} & \textbf{0.63} \\
    BCP &{}&{} &82.91 & 70.97 & 6.43 & 2.25 \\
    Co-BioNet &{}&{}& 84.01 & 70.00 & 5.35 & 2.75 \\
    \hline
    EPCL-EN&\multirow{2}*{12(20\%)} &\multirow{2}*{50(80\%)} & 83.79 & 72.53 & 5.86 & 2.64\\
    EPCL-JUQ&{}&{} & \textbf{84.45} & \textbf{73.61} & 5.12 & 2.43\\
    \hline
  \end{tabular}}
  \label{Pancreas}
\end{table}
\begin{table}[htbp]
    \renewcommand{\arraystretch}{1.3}
	\label{table5}
	\centering
        \caption{Comparison with state-of-the-art models on Aortic Dissection dataset}
			\setlength{\tabcolsep}{1.5mm}{
                \begin{tabular}{c|c c|c c c c}
				\hline
                    \multicolumn{1}{c|}{\multirow{2}*{Model}}& \multicolumn{2}{c|}{Scans Used} &\multicolumn{4}{c}{Metrics}\\\cline{2-7}
				&{Labeled} & {Unlabled} &Dice$\uparrow$&Jaccard$\uparrow$&95HD$\downarrow$&ASD$\downarrow$  \\\hline
				MT&\multirow{5}*{(20\%)} & \multirow{5}*{(80\%)} &  53.78& 38.54& 7.49& 1.87 \\
				UA-MT&&&65.78& 51.20& 6.19& 1.60  \\
				FUSSNet&&& 72.53& 59.52& 5.67& 1.77  \\
				URPC&&& 75.50& 63.68& 6.77& 1.02  \\
				UPCoL&&&  76.19&  64.45& 4.82& 1.33 \\ \hline
			EPCL-EN&(20\%)&(80\%)&\textbf{79.94} & \textbf{68.32}&\textbf{3.63} & \textbf{0.74}\\
            EPCL-JUQ&(20\%)&(80\%)&\textbf{80.89} & \textbf{69.14}&\textbf{3.45} & \textbf{0.71} \\
            \hline
		\end{tabular}}
 \label{1}
\end{table}
To better demonstrate the effectiveness of the proposed EPCL-JUDA framework, we visualize its predictions on the LA dataset and compare them with URPC' predictions and ground truth labels in Fig. \ref{vis}. The results show that EPCL-JUDA's predictions are much closer to the actual labels than the baseline, with significant improvements in Dice and Jaccard scores. Specifically, for the first and third examples, the baseline predictions miss sections that EPCL-JUDA accurately captures. 

\subsection{Ablation studies}
We conduct extensive ablation studies to explore the effectiveness of each component of the proposed framework and various optimal parameter settings. 
\begin{table}[htbp]\small
\renewcommand{\arraystretch}{1.2}
\centering
\caption{Ablation studies of different components in the proposed framework of EPCL-JUDA on the LA dataset. $w/o$ JUQ denotes that the joint uncertainty quantification is replaced as entropy uncertainty measurement. $w/o$ DA represents that the additional augmented labeled and unlabeled data is removed. $w/o$ PRO denotes removing the prototype network and directly utilizing features of decoders to generate prototypes.}
  \setlength{\tabcolsep}{0.8mm}{\begin{tabular}{c|cc|cccc}
    \hline
    \multirow{2}*{Method} &\multicolumn{2}{c|}{Scans Used} &\multicolumn{4}{c}{Metrics}\\
    \cline{2-7}
    {}&{Labeled} & {Unlabled} & Dice$\uparrow$ & Jaccard$\uparrow$ & 95HD$\downarrow$ & ASD$\downarrow$ \\
    \hline
    EPCL-JUDA & 8(20\%) & 0 & \textbf{92.24} & \textbf{85.57} & \textbf{4.63}& \textbf{1.47}\\
    \hline
    $w/o$ JUQ & 8(20\%) & 0 & 91.88 & 84.86 & 4.78 & 1.66 \\
    $w/o$ DA & 8(20\%) & 0 & 91.38 & 84.22 & 6.59 & 2.34 \\
    $w/o$ PRO & 8(20\%) & 0 & 91.94 & 85.02 & 4.77 & 1.63 \\
    \hline
  \end{tabular}}
  \label{component}
\end{table}
Firstly, we attempt to explore the effects of different components within the EPCL-JUDA framework. These components include Joint Uncertainty Quantification (JUQ), Data Augmentation (DA), and the Prototype Network (PRO). Table \ref{component} presents the experimental results on the LA dataset. As shown in the table, removing JUQ ($w/o$ JUQ) and utilizing entropy as uncertainty measurement result in a slight performance drop, indicating that the introduction of mean and variance uncertainty can further enhance the model's sensitivity to anomalous data and improve its robustness, thereby further enhancing the model's segmentation performance. The removal of Data Augmentation ($w/o$ DA) leads to a significant decrease in performance metrics, underscoring the value of augmented labeled and unlabeled data for improving the quality of global prototypes that are utilized to conduct consistency learning. Additionally, removing the Prototype Network ($w/o$ PRO) and directly utilizing features of decoders to generate prototypes only results in a relatively small performance drop. However, it also demonstrates that the former not only performs better than the latter but also significantly reduces the model's memory requirements. Overall, the full EPCL-JUDA method, which integrates all these components, achieves the best performance across all metrics, confirming the necessity of each component in the proposed EPCL-JUDA framework.\\
\textbf{Analysis of Memory Requirements.} 
As mentioned earlier, directly upsampling features of each decoder layer to generate prototype features results in high memory requirements. Since decoder features generally have numerous channels, processing all high-dimensional features at once will result in significant memory consumption. To address this issue, we design a prototype network to generate prototype features, effectively reducing the memory needed by the model. Table \ref{memory} demonstrates the memory usage of two different generation techniques for prototype features, which validate the efficiency of the prototype network.
\begin{table}[htbp]
    \centering
    \caption{Memory Usage Comparison: Decoder Features vs Prototype Network}
    \begin{tabular}{cccc}
    \toprule
    \textbf{Decoder Layer} & 1 & 2 & 3\\
    \midrule
    \textbf{Memory Usage (GB) - Decoder Features} & 18.65 & 15.93 & 14.51 \\
    \textbf{Memory Usage (GB) - Prototype Network} & 4.21 & 6.54 &  6.65\\
    \hline
\end{tabular}
\label{memory}
\end{table}
It can be observed that using prototype network significantly reduces the memory usage compared to directly using the decoder's features. For example, in the first decoder layer, the memory usage drops from 18.65 GB to 4.21 GB, demonstrating a substantial reduction in memory consumption. This trend is consistent across all decoder layers, indicating the effectiveness of the prototype network in memory usage. The reason why prototype network essentially maintains low memory consumption is that regardless of which decoder layer's features are being processed, the dimension of final upsampled features is always equal to the number of classes, maintaining stable memory requirements.\\
\textbf{Combination pattern analysis of unlabeled data.} 
We investigate effective methods for combining unlabeled data with augmented unlabeled data. The approaches are categorized into four types: 1) concatenate the data and feed it into the teacher network; 2) feed the data separately into the teacher network and construct multiple unlabeled prototypes consistency learning ; 3) use the reliability map from augmented data to enhance the original unlabeled data training; 4) use the reliability map from original data to optimize the augmented data training in the teacher network.
\begin{table}[htbp]\small
\renewcommand{\arraystretch}{1.2}
\centering
\caption{Ablation experiments analyzing the combination methods of original labeled data and augmented unlabeled data}
  \setlength{\tabcolsep}{0.8mm}{\begin{tabular}{c|cc|cccc}
    \hline
    \multirow{2}*{Method} &\multicolumn{2}{c|}{Scans Used} &\multicolumn{4}{c}{Metrics}\\
    \cline{2-7}
    {}&{Labeled} & {Unlabled} & Dice$\uparrow$ & Jaccard$\uparrow$ & 95HD$\downarrow$ & ASD$\downarrow$ \\
    \hline
    1) & 8(20\%) & 0 & 92.03 & 85.28 & 4.73 & 1.66 \\
    2) & 8(20\%) & 0 & \textbf{92.24} & \textbf{85.57} & 4.63 & \textbf{1.47} \\
    3) & 8(20\%) & 0 & 91.99 & 85.19 & \textbf{4.52} & 1.50\\
    4) & 8(20\%) & 0 & 91.88 & 85.01 & 4.88 & 1.68\\
    \hline
  \end{tabular}}
  \label{con}
\end{table}
The experimental results shown in Table \ref{con} demonstrate that 2) provides the best performance, highlighting the importance of processing original and augmented data separately and leveraging consistency learning with multiple unlabeled prototypes. Furthermore, the performance of these different combination methods do not show significant variations, indicating that regardless of the combination method used, the augmented unlabeled data consistently provides a substantial amount of valuable semantic information, thereby enhancing the model's performance.\\
\begin{figure}
    \centering
    \includegraphics[scale=0.28]{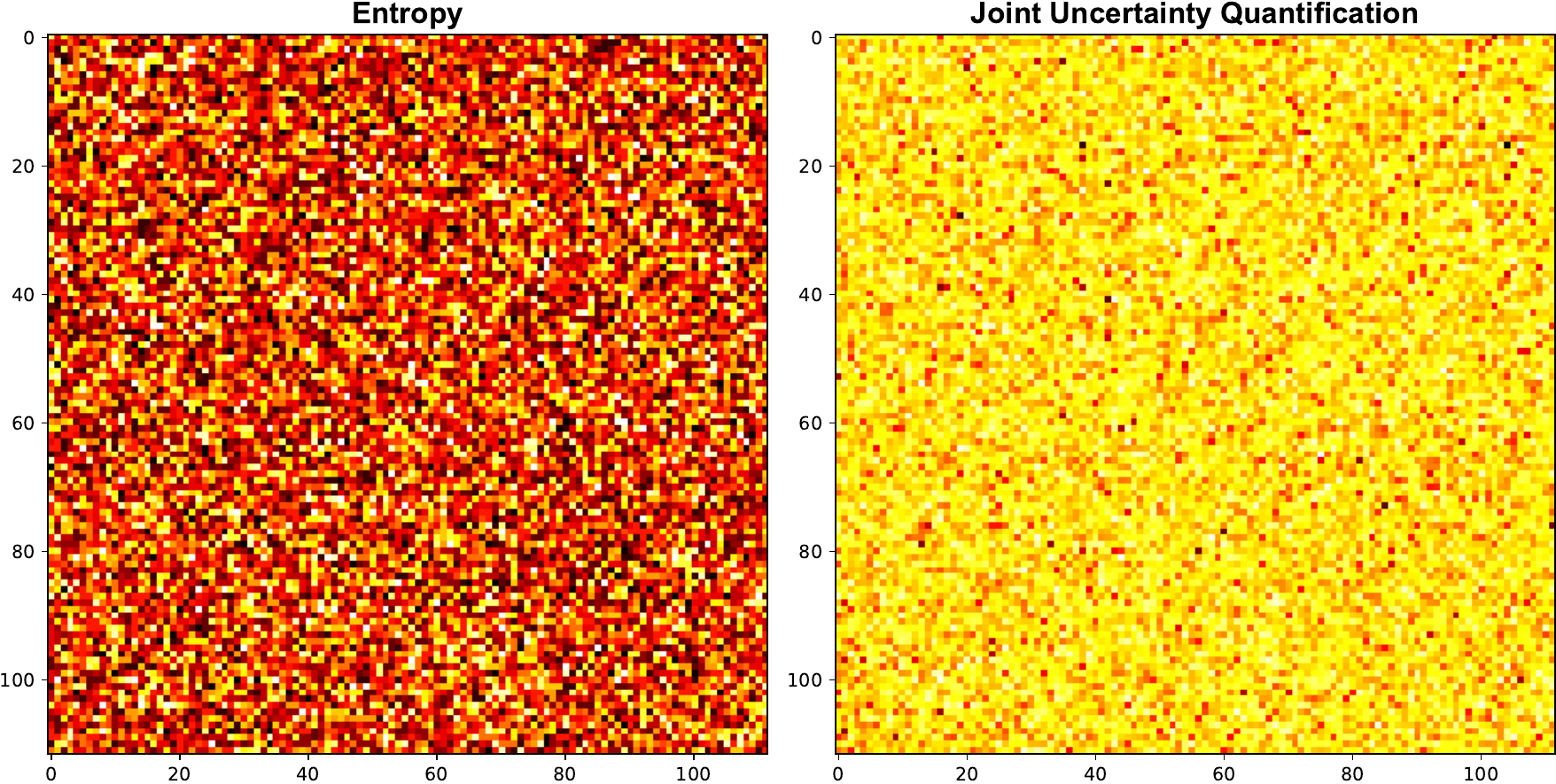}
    \caption{The reliability map generated by Entropy and JUQ}
    \label{uncertainty}
\end{figure}
\textbf{Ablation for Joint Uncertainty Quantification.}
To validate the effectiveness of the proposed JUA method, we randomly initialize multiple predictions and obtain their reliability maps using entropy and JUA. Fig. \ref{uncertainty}. shows the visualization results of a slice in reliability maps obtained by two methods, where the yellow areas indicate high certainty. The black areas demonstrate that entropy uncertainty is sensitive to noise, especially when dealing with augmented data, which can easily cause over-sensitivity. In contrast, JUA, by comprehensively considering both entropy and distributional uncertainty, improves robustness against noise. The figure shows that the high certainty regions of JUA are more stable and broader.

\section{Conclusions}
This paper presents a novel framework, Efficient Prototype Consistency Learning via Joint Uncertainty Quantification and Data Augmentation, for semi-supervised medical image segmentation. The framework addresses the limitations of previous prototype-based methods by generating high-quality prototypes with the integration of joint uncertainty quantification and data augmentation based on a Mean-Teacher structure. Besides, a prototype network is devised to significantly reduce memory consumption and further improve segmentation performance. Extensive experiments on public benchmark datasets including the Left Atrium, Pancreas-NIH, and Aortic Dissection demonstrate that the proposed EPCL-JUA framework consistently outperforms previous SOTA methods, achieving superior segmentation performance.

\bibliographystyle{splncs04}
\bibliography{IEEEfull.bib}
\end{document}